\documentclass{article}
\usepackage[preprint]{neurips_2026}

\usepackage[utf8]{inputenc} 
\usepackage[T1]{fontenc}    
\usepackage{hyperref}       
\usepackage{url}            
\usepackage{booktabs}       
\usepackage{graphicx}       
\usepackage{amsmath}        
\usepackage{amsfonts}       
\usepackage{nicefrac}       
\usepackage{microtype}      
\usepackage[dvipsnames]{xcolor} 
\usepackage{pifont}         
\usepackage{multirow}
\usepackage{placeins}       
\usepackage{float}          
\usepackage{algorithm}
\usepackage{algpseudocode}  

\newcommand{\cmark}{\textcolor{OliveGreen}{\ding{51}}}
\newcommand{\xmark}{\textcolor{BrickRed}{\ding{55}}}
\newcommand{\pmark}{\textcolor{Goldenrod}{$\sim$}}

\title{LePlanner: An Iterative Amortized Controller For World Models}

\author{%
  \begin{minipage}[t]{0.95\textwidth}
    \centering
    \textbf{Saksham Bansal\textsuperscript{1*}, Om Naphade\textsuperscript{2*}, Chayan Aggarwal\textsuperscript{3*}, Vrishin M\textsuperscript{4}}\\[6pt]
    \normalfont\small
    \textsuperscript{1}Department of Mechanical Engineering,
    \textsuperscript{2}Department of Physics,
    \textsuperscript{3}Department of Mathematics,
    \textsuperscript{4}Department of Chemical Engineering\\
    {\fontsize{10pt}{12pt}\selectfont\ Indian Institute of Technology Roorkee}\\[2pt]
    {\fontsize{10pt}{2pt}\selectfont\{saksham\_b@me, om\_nn@ph, chayan\_a@ma, vrishin\_m@ch\}.iitr.ac.in}\\[3pt]
  \end{minipage}
}

\begin{document}

\raggedbottom
\renewcommand{\topfraction}{0.9}
\renewcommand{\bottomfraction}{0.8}
\renewcommand{\textfraction}{0.07}
\renewcommand{\floatpagefraction}{0.7}
\maketitle

\begingroup
\renewcommand{\thefootnote}{\fnsymbol{footnote}}
\footnotetext[1]{Equal contribution.}
\renewcommand{\thefootnote}{}
\footnotetext[0]{Models: \url{https://huggingface.co/nottygian/Leplanner}.}
\endgroup
\begin{abstract}
    World models trained with joint-embedding predictive architectures learn compact, structured latent manifolds from physical interaction, yet planning in those latents still relies on one of two costly families of methods. Search-based planners (CEM, MPPI, iCEM) optimize action sequences through many predictor rollouts, achieving strong success at the price of high per-decision compute and latency. Policy-based methods (behavior cloning, GC-IDM) amortize inference into a single forward pass, but degrade on contact-rich tasks where the demonstration distribution is multi-modal. Neither family offers both low latency and robust goal-reaching in a learned latent space. We propose LePlanner, an amortized iterative controller that refines latent action sequences to reach a goal state, trained with an arrival--hold loss that rewards reaching the goal at the earliest feasible horizon and staying there, plus an action gaussian loss which keeps actions in their true manifold. Across goal-directed environments spanning navigation, contact-rich manipulation, and continuous control, LePlanner matches or exceeds the success of test-time search planners while requiring an order-of-magnitude fewer predictor invocations and $3$--$49\times$ lower wall-clock per decision. We report $98\%$ on PushT, $100\%$ on Reacher, $100\%$ on TwoRooms, and $92\%$ on the ogbench cube. The results show that, in a sufficiently regularized latent space, much of the structure that search discovers online can be amortized into a lightweight learned iterative policy, yielding fast, horizon-aware, nonlinear physical control without online optimization.
\end{abstract}

\section{Introduction}
\label{sec:introduction}

A central goal of artificial intelligence is to develop agents that learn useful behavior across diverse tasks and environments directly from observations \citep{levine2016visuomotor}.
World models are one promising approach: they learn to predict how an environment state evolves in response to an action, so agents can evaluate possible action sequences before executing them, effectively planning inside a learned model of the world \citep{ha2018worldmodels,hafner2019planet}.
Joint Embedding Predictive Architectures (JEPAs) learn such models in a compact latent space rather than predicting every detail of future observations, which works particularly well when the observation channel is high dimensional \citep{lecun2022path,assran2023ijepa,sobal2022slowfeatures}.
Recent work on LeWorldModel (LeWM) demonstrated that a JEPA based world model can be trained stably from raw pixels while supporting efficient visual control \citep{maes2026leworldmodel}.

Planning through a learned world model forces a three-way choice: search-based planners (ie.\ CEM) are accurate but costly and high-latency, solving a full optimization loop at every decision \citep{rubinstein1999cem,chua2018pets,hafner2019planet}; cloning-based policies (GC-IDM) amortize inference into a single forward pass but degrade on contact-rich and nonlinear tasks where the demonstration distribution is poorly conditioned \citep{nguyen2026latentgeometrysearchamortizing}; and model-based policy methods (TDMPC2) train the controller inside the world model's imagination, coupling the policy to a task-specific dynamics model that demands substantial compute to train on real environments and can struggle with complex long-horizon structure \citep{hansen2024tdmpc2scalablerobustworld}.

We also find that a conventional terminal objective used in such controllers creates a rather subtle failure when deployed in a varied receding-horizon setting \citep{mayne2000mpc}.
The objective trains the controller to reach the goal at the end of its planning horizon.
When deployed, only the first part of the plan is executed before replanning, which resets the terminal deadline, so the controller repeatedly approaches the goal while postponing actual arrival in many cases.
We refer to this behavior as \emph{horizon-reset procrastination}.

To resolve this mismatch and to bring the cost of non-amortized controllers down, we propose LePlanner, an amortized controller that learns to construct and iteratively refine action plans through a frozen LeWM predictor \citep{srinivas2018upn,marino2018iterative}, trained with an \emph{arrival-and-hold} objective.
During training, the controller is supervised to reach the goal at the temporal offset from which that goal was sampled, rather than at the fixed end of every plan, and is then encouraged to remain near the goal for the rest of the predicted trajectory.
The temporal offset is used only to define the training loss and is never provided to the controller at inference, so the method requires no modification to the frozen world model or controller architecture.
In the currently recorded PushT evaluation, LePlanner reaches \textbf{$98\%$ success} while replanning after every action block: the corrected objective removes the procrastination behavior and allows the controller to use frequent replanning without repeatedly deferring the goal.
LePlanner also achieves high success rate in evaluations such as TwoRoom and Reacher.

\begin{figure}[tbp]
  \centering
  \includegraphics[width=0.85\linewidth]{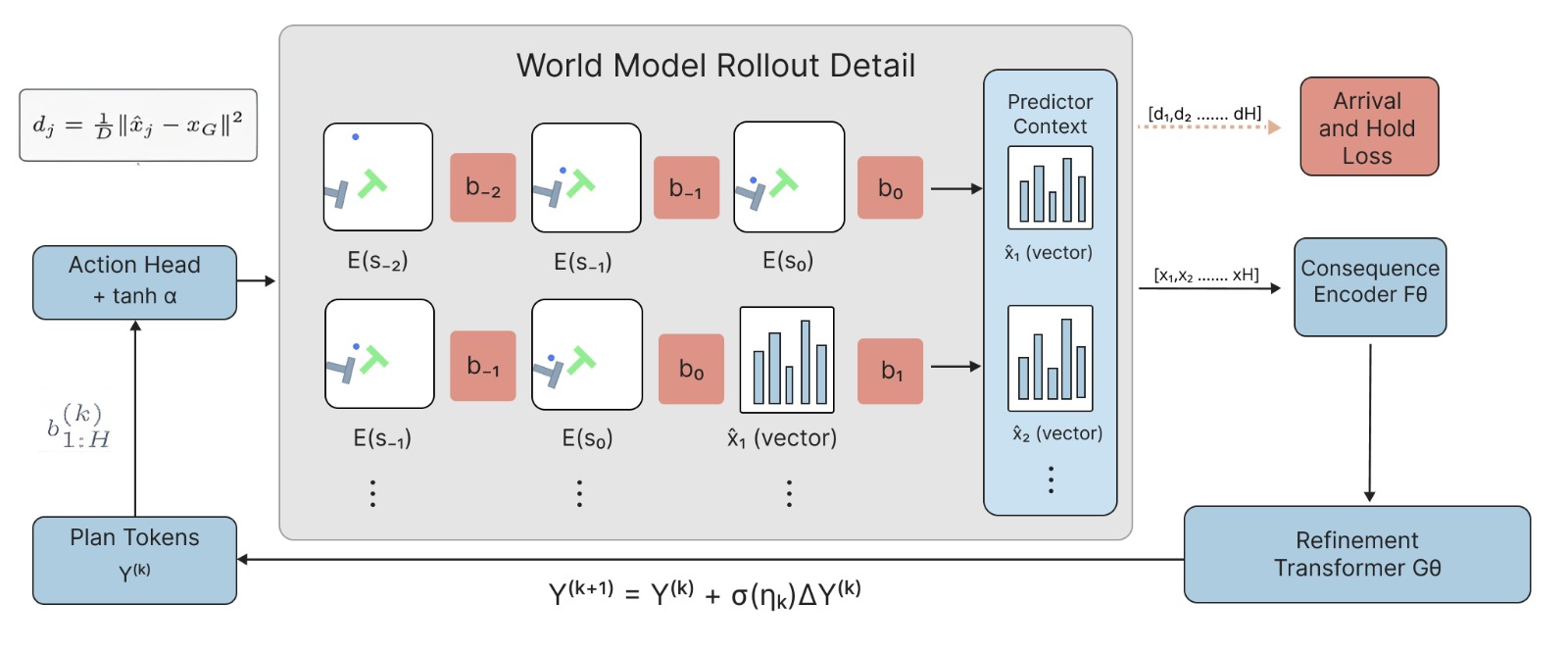}
\caption{LePlanner architecture. The frozen JEPA predictor (blue) rolls out imagined latents $\bar x_1,\bar x_2,\ldots$ from context $(E(s_i)+b_i)$, where $E(s_i)$ is the latent embedding of state $s_i$ and $b_i$ is the executed action. $Y^{(k)}$ denotes the action plan at refinement step $k$, produced by the Refinement Transformer and decoded into real-environment action states by the action head $h_\theta$.}

  \label{fig:overview}
\end{figure}

\FloatBarrier

\section{Related Work}

\paragraph{Latent world models and visual planning.}
World models learn predictive representations that allow agents to reason about the
consequences of their actions before acting \citep{ha2018worldmodels}.
Early latent world-model approaches either optimized action sequences online, as in
PlaNet \citep{hafner2019planet}, or jointly learned task-oriented latent dynamics
and terminal values for model-predictive control \citep{hansen2022tdmpc}.
Joint Embedding Predictive Architectures instead predict future latent
representations rather than reconstructing observations
\citep{lecun2022path,assran2023ijepa}, and DINO-based models show that
self-supervised vision features yield smooth latent spaces suitable for
downstream control \citep{caron2021dino,oquab2023dinov2}.
This principle has been extended to action-conditioned latent dynamics
\citep{sobal2022slowfeatures} and to planning over frozen visual
representations \citep{zhou2025dinowm}.
LeWorldModel learns an action-conditioned JEPA end-to-end from pixels and
provides the frozen world model used in our work \citep{maes2026leworldmodel}.
Recent studies improve latent planning by reducing the mismatch between
world-model training and planning \citep{parthasarathy2025traintest}, or by
regularizing latent trajectories to improve the geometry of the planning
objective \citep{wang2026straightening}.
Our work is complementary: we keep the learned representation and dynamics
fixed and study how the controller's training objective interacts with
receding-horizon execution.

\paragraph{Planning in latent worlds: search, cloning, and learned policies.}
Planning over a learned world model is most often done by test-time search.
The cross-entropy method (CEM) \citep{rubinstein1999cem} and
MPPI \citep{williams2017mppi} optimize action sequences by iteratively
refining a distribution over candidate plans through the learned dynamics, and
PETS \citep{chua2018pets} combines such sampling with probabilistic
ensembles; iCEM adds implicit differentiation for continuous refinement.
These search planners achieve strong success but repeat substantial computation
at each decision, evaluating hundreds to thousands of latent rollouts per step.
An alternative is to amortize planning into a learned policy: goal-conditioned
behavior cloning (GCBC) and Goal-Conditioned Inverse Dynamics Models (GC-IDM)
map a current latent state, goal latent, and horizon directly to the next action
\citep{nguyen2026latentgeometrysearchamortizing}.
Because they are trained purely on demonstration data, such policies amortize
inference into a single forward pass but degrade on contact-rich or multi-modal
tasks where the behavioral cloning distribution is poorly conditioned.
A third family trains policies inside the world model's imagination: Dreamer
\citep{hafner2020dreamer,hafner2023dreamerv3} trains an actor-critic entirely on
imagined latent rollouts, and TDMPC2 \citep{hansen2024tdmpc2scalablerobustworld}
extends this with joint latent-reward learning and TD-planning over short
imagination horizons.
Our controller bridges these families. Like search planners it reasons
through the learned dynamics over multiple horizons; like amortized
policies it performs a fixed number of learned refinements at deployment;
and like model-based policy methods it trains the controller end-to-end
-- but over a \emph{frozen, task-agnostic} world model, with an explicit
arrival-and-hold objective.

\paragraph{Amortized and iterative inference.}
Iterative amortized inference combines a learned initial prediction with
repeated learned corrections \citep{marino2018iterative}, and Universal Planning
Networks \citep{srinivas2018upn} learn representations in which plans are
optimized through differentiable latent rollouts.
More broadly, iterative refinement underlies energy-based
models \citep{lecun2006energybased} and consistency models
\citep{song2023consistency}.
Our controller follows this amortized formulation, but unlike online CEM the
search procedure is learned during training, and unlike Universal Planning
Networks the world model is pretrained and frozen and the controller is not
trained by imitating expert action sequences. This separation lets us change the
controller objective without changing the underlying visual representation
or dynamics.

\section{Method: LePlanner}
\label{sec:method}

\begin{figure}[tbp]
\centering \includegraphics[width=\linewidth]{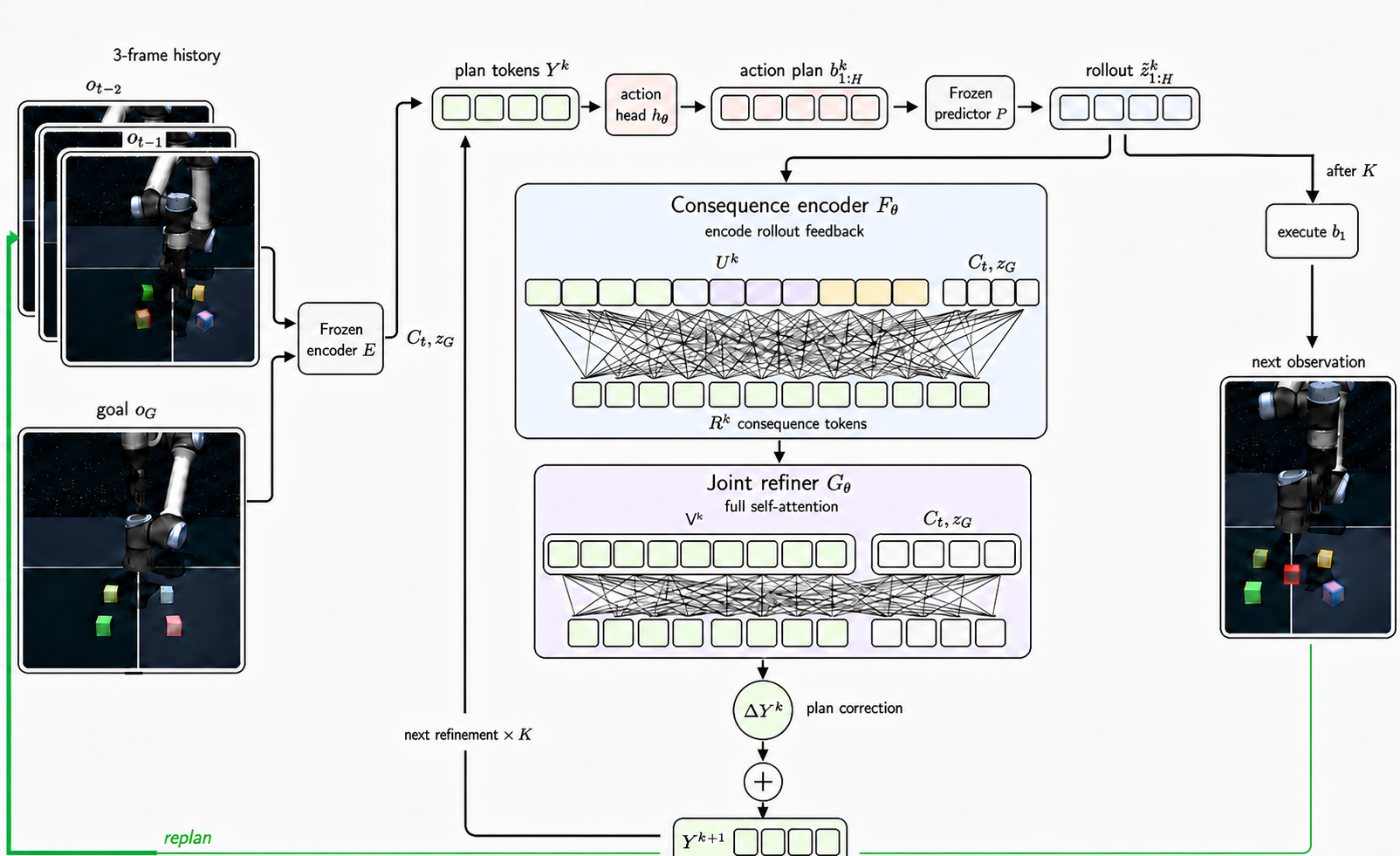}
\caption{\textbf{The LePlanner architecture.} From a 3-frame observation history and a goal image $o_G$, the frozen encoder $E$ produces context tokens $C_t$ and goal latent $z_G$. Plan tokens $Y^k$ are decoded by the action head $h_\theta$ into an action plan $b^k_{1:H}$, which the frozen predictor $P$ rolls out into imagined latents $\tilde{z}^k_{1:H}$. The consequence encoder $F_\theta$ attends jointly over $Y^k$, the rollout $\tilde{z}^k$, the rollout-to-goal residual $\tilde{z}^k - z_G$, and the predicted distance-to-goal $d^k$ with contexts and goal to produce consequence tokens $R^k$. The joint refiner $G_\theta$ then applies full self-attention jointly over $Y^k$ and $R^k$, with the conditioning tokens $C_t, z_G$ to predict a plan correction $\Delta Y^k$, yielding $Y^{k+1}$. This repeats for $K$ refinements; after the final iteration the first action block $b_1$ is executed, a new observation is received, and the controller replans.}
\label{fig:leplanner-architecture}
\end{figure}

We consider offline, visual, goal-conditioned control using a pretrained and frozen latent world model.
Our objective is to learn a controller that produces action plans from an observation history and a goal image, without imitating the actions in the offline dataset.

\subsection{Problem setting}
\label{sec:problem-setting}

Let an offline trajectory be $\tau=(o_0,a_0,o_1,a_1,\ldots)$, where $o_t$ is a visual observation and $a_t\in\mathbb{R}^{d_a}$ is an environment action.
We use a pretrained LeWorldModel (LeWM) \citep{maes2026leworldmodel}, consisting of an image encoder $E:o_t\mapsto z_t\in\mathbb{R}^{D}$ and an action-conditioned latent predictor $P$.
Environment actions are grouped into blocks of $f$ consecutive actions; a block is denoted by $b_j\in\mathbb{R}^{A}$, where $A=f d_a$.
At time $t$, the controller receives $N$ context latents $C_t=(z_{t-N+1},\ldots,z_t)$, the $N-1$ action blocks connecting these observations, and a goal latent $z_G=E(o_G)$.
It produces a plan $b_{1:H}=(b_1,\ldots,b_H)$ of $H$ future action blocks.
Given a candidate plan, the frozen predictor generates an autoregressive latent rollout:
\begin{equation}
    \hat z_{1:H}
    =
    \mathcal{R}_{P}
    \left(C_t, b_{t-N+1:t-1}, b_{1:H}\right),
    \label{eq:latent-rollout}
\end{equation}
where $\mathcal{R}_{P}$ applies $P$ using the same sliding observation and action history used during world-model training.
We measure the predicted distance to the goal after block $j$ as $d_j = \tfrac{1}{D}\lVert \hat z_j-z_G\rVert_2^2$.
The parameters of $E$ and $P$ remain frozen throughout controller training.
The cached outputs of $E$ are treated as data, while gradients are retained through the operations of $P$ so that the controller can learn how its proposed actions affect the predicted trajectory.

\subsection{LePlanner}
\label{sec:leplanner}

LePlanner amortizes trajectory optimization into a fixed number of learned refinement steps.
It maintains one hidden plan token $y_j\in\mathbb{R}^{W}$ for each future action block, $Y^{(k)}=(y_1^{(k)},\ldots,y_H^{(k)})$, where $k$ denotes the refinement iteration.
The context latents and goal latent are projected into the controller width and augmented with learned positional and role embeddings; these conditioning tokens are read by every refinement step.
The initial plan is produced from learned plan queries, $Y^{(0)} = G_{\theta}([Q;\mathbf{0}],C_t,z_G)$, where $Q$ contains one learned query per plan position and the zero argument indicates that no predicted consequences are available for the initial proposal.
An action head $h_\theta$ decodes every plan token into a bounded action block, $b_j^{(k)} = c_a+s_a\odot\tanh(h_\theta(y_j^{(k)}))$, where $c_a$ and $s_a$ transform the raw action bounds into the normalized action space used by LeWM, so the $\tanh$ parameterization enforces the environment action bounds without clipping.

For each refinement, the decoded plan is rolled through the frozen predictor using Equation~\eqref{eq:latent-rollout}.
For plan position $j$, LePlanner constructs the consequence feature
$u_j^{(k)} = [,y_j^{(k)};\ \hat z_j^{(k)};\ \hat z_j^{(k)}-z_G;\ d_j^{(k)},]$.
A consequence encoder $F_\theta$ interprets these features jointly with the observation and goal conditioning, and a refinement transformer $G_\theta$ then predicts a correction for all plan tokens, which updates the plan through a learned step size:
\begin{equation}
\begin{aligned}
R^{(k)}
&= F_\theta\!\left(U^{(k)},C_t,z_G\right),
&\qquad
V^{(k)}
&= \left[Y^{(k)};R^{(k)}\right], \\
\Delta Y^{(k)}
&= G_\theta\!\left(V^{(k)},C_t\right),
&
Y^{(k+1)}
&= Y^{(k)}+\sigma(\eta_k)\Delta Y^{(k)}.
\end{aligned}
\label{eq:refinement}
\end{equation}
for $k=0,\ldots,K-1$, where $V^{(k)}$ is the concatenation of the current plan and predicted consequences, and $\sigma(\eta_k)\in(0,1)$ and one scalar step size is learned for each refinement.
Both $F_\theta$ and $G_\theta$ are shared across all $K$ iterations, so increasing the refinement depth increases inference computation but does not introduce new parameters.
The refinement transformer ($G_\theta$) applies full self-attention across all $H$ plan positions before any action block is committed, allowing early and late parts of the plan to be revised jointly.

\subsection{Arrival-and-hold objective}
\label{sec:arrival-hold}

A training example is constructed by sampling a future observation from the same offline trajectory as the current context.
Let $q\in\{1,\ldots,H\}$ be the number of world-model transitions between the current observation and this hindsight-relabeled goal \citep{andrychowicz2017her}.
The dataset therefore provides not only the goal latent $z_G$, but also the offset at which that goal occurs.
A fixed-terminal objective would always penalize $d_H$, regardless of the sampled offset $q$; under receding-horizon execution, this attaches arrival to the end of the repeatedly reset plan.
Instead, LePlanner evaluates the predicted trajectory at the goal's data-derived offset:
\begin{equation}
    \mathcal{L}^{(k)}_{\mathrm{AH}}
    =
    \underbrace{d_q^{(k)}}_{\text{arrival}}
    +
    \lambda_h
    \underbrace{
        \frac{1}{H-q}
        \sum_{j=q+1}^{H}d_j^{(k)}
    }_{\text{hold}},
    \label{eq:arrival-hold}
\end{equation}
where the hold term is defined as zero when $q=H$.
The arrival term asks the predicted plan to reach the goal at the temporal offset from which the goal was sampled, and the hold term discourages trajectories that briefly pass through the goal and then move away.
Importantly, $q$ is used only to index the training loss; it is never included in the controller input.
At inference, LePlanner receives only the observation history and goal, and must infer an appropriate plan without access to a deadline.

\subsection{Deep supervision across refinements}
\label{sec:deep-supervision}

LePlanner produces an initial plan followed by $K$ refined plans, and we supervise every one of these $K+1$ predictions:
\begin{equation}
    \mathcal{L}_{\mathrm{ref}}
    =
    \frac{
        \sum_{k=0}^{K}
        \rho_k\mathcal{L}^{(k)}_{\mathrm{AH}}
    }{
        \sum_{k=0}^{K}\rho_k
    },
    \qquad
    \rho_k=2^k.
    \label{eq:deep-supervision}
\end{equation}
Later refinements receive larger weights, while every intermediate plan retains a direct gradient signal.
This makes the initial prediction useful on its own and trains each application of the shared refinement operator rather than supervising only its final output.

\subsection{Constraining plans to offline support}
\label{sec:support}

Optimizing actions through a learned world model can exploit inaccuracies outside the action distribution represented in the offline data \citep{fujimoto2019bcq,kumar2019bear,yu2020mopo,kidambi2020morel}.
We therefore fit a conditional Gaussian mixture density $\beta_\psi(b\mid C)$ to real action blocks and their latent contexts, trained separately and frozen before LePlanner training.
For a candidate action block, we define the normalized negative log density $r(C,b) = -\tfrac{1}{A}\log\beta_\psi(b\mid C)$, and let $c_{95}$ be the $95$th percentile of this score on real action blocks.
For the refined plans, we apply the one-sided support penalty
\begin{equation}
    \mathcal{L}_{\mathrm{sup}}
    =
    \frac{1}{KH}
    \sum_{k=1}^{K}
    \sum_{j=1}^{H}
    \left[
        \max\left(
            0,
            r(C_j^{(k)},b_j^{(k)})-c_{95}
        \right)
    \right]^2,
    \label{eq:support-loss}
\end{equation}
where $C_j^{(k)}$ is the imagined context preceding block $j$.
The penalty is zero for action blocks inside the selected support region, so it does not train LePlanner to reproduce a particular demonstrated action; the dataset actions are used only to fit $\beta_\psi$ and never serve as controller targets.
The complete controller objective is $\mathcal{L}_{\mathrm{LePlanner}} = \mathcal{L}_{\mathrm{ref}} + \lambda_{\mathrm{sup}}\mathcal{L}_{\mathrm{sup}}$.
Only the parameters of LePlanner are optimized; the LeWM encoder, LeWM predictor, and behavior-density model remain frozen.

\subsection{Receding-horizon execution}
\label{sec:inference}

At deployment, LePlanner encodes the current observation history and goal, constructs its initial plan, and performs $K$ learned refinements.
No gradient computation or online parameter update is required, and the final plan $b_{1:H}^{(K)}$ is returned to the environment policy.
We use receding-horizon execution: the policy executes a prefix of the plan, observes the resulting state, and invokes LePlanner again \citep{mayne2000mpc}.
The main closed-loop configuration executes one action block before replanning.
Thus, the world model remains inside the inference loop, but population-based trajectory search is replaced by a fixed number of amortized refinements.

\begin{figure}[H]
  \centering
  \includegraphics[width=0.75\linewidth]{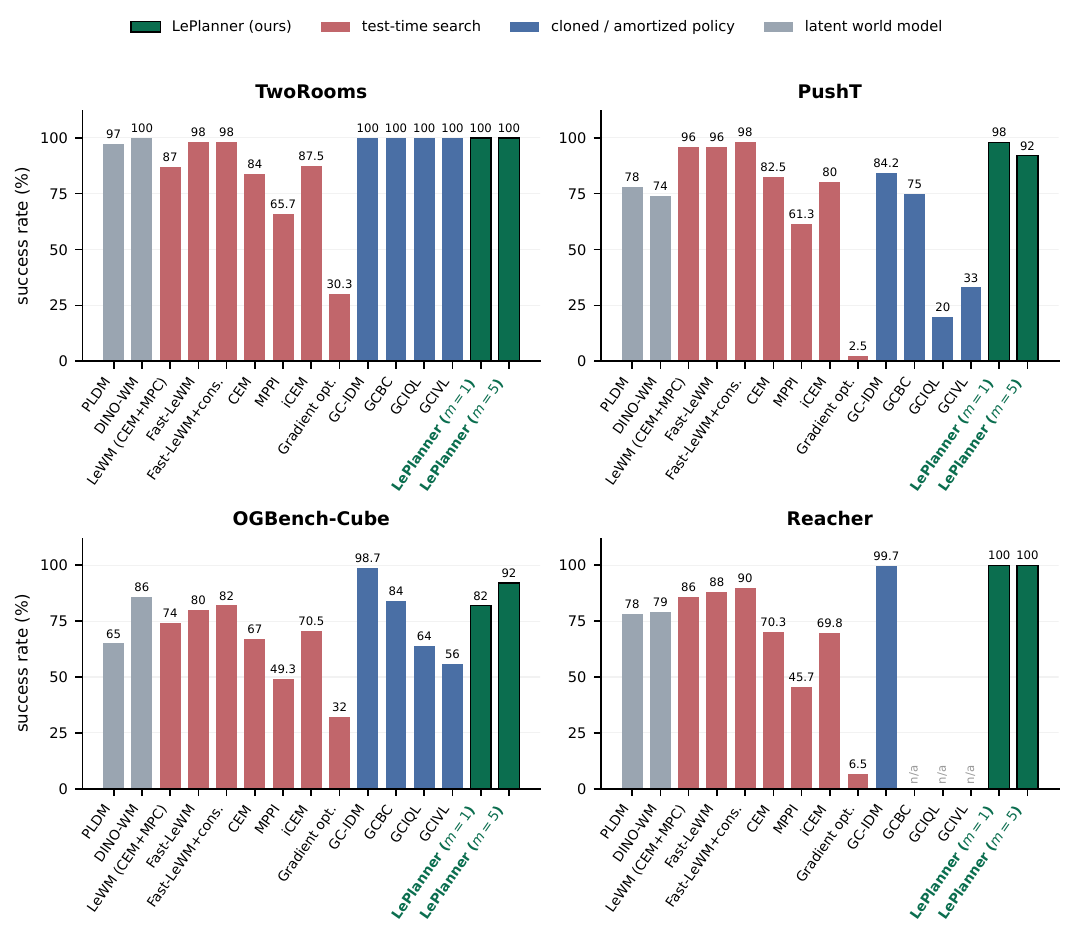}
  \caption{\textbf{Success across all compared methods.} One panel per environment; bars are coloured by method family. LePlanner (green, outlined) is shown at both receding horizons.}
  \label{fig:success}
\end{figure}

\section{Results}
\label{sec:results}

We evaluate LePlanner against three families of latent planners: test-time
search (CEM, MPPI, iCEM, gradient optimisation), amortised policies trained by
goal-conditioned cloning (GCBC, GCIQL, GCIVL, GC-IDM), and latent world models
evaluated with their own planners (PLDM, DINO-WM, LeWM, Fast-LeWM).
The four benchmark environments are PushT \citep{chi2023diffusionpolicy},
Reacher, TwoRooms, and the OGBench cube task \citep{park2025ogbench}.

\subsection{Goal-reaching success}
\label{sec:res-success}

Figure~\ref{fig:success} reports success on the four LeWM benchmark
environments. LePlanner reaches $100\%$ on TwoRooms, $94\%$ on PushT, $100\%$ on
Reacher, and $92\%$ on OGBench-Cube. Against the matched CEM baseline run under
an identical protocol, it improves success on every environment we measured.
Two properties are worth stating before the numbers are read. First, LePlanner
is close to flat in $m$, the number of action blocks executed before replanning
($100 \rightarrow 100$ on TwoRooms, $94 \rightarrow 88$ on PushT, $100
\rightarrow 100$ on Reacher, $82 \rightarrow 92$ on Cube), the signature of a
time-consistent closed-loop policy rather than a good open-loop one. Second, CEM
is not flat: on PushT it moves $34 \rightarrow 90$ between $m{=}1$ and $m{=}5$,
so a single receding horizon cannot summarise it, and we report both throughout.

\subsection{The cost of a decision}
\label{sec:res-cost}

\begin{figure}[H]
  \centering
  \includegraphics[width=.86\linewidth]{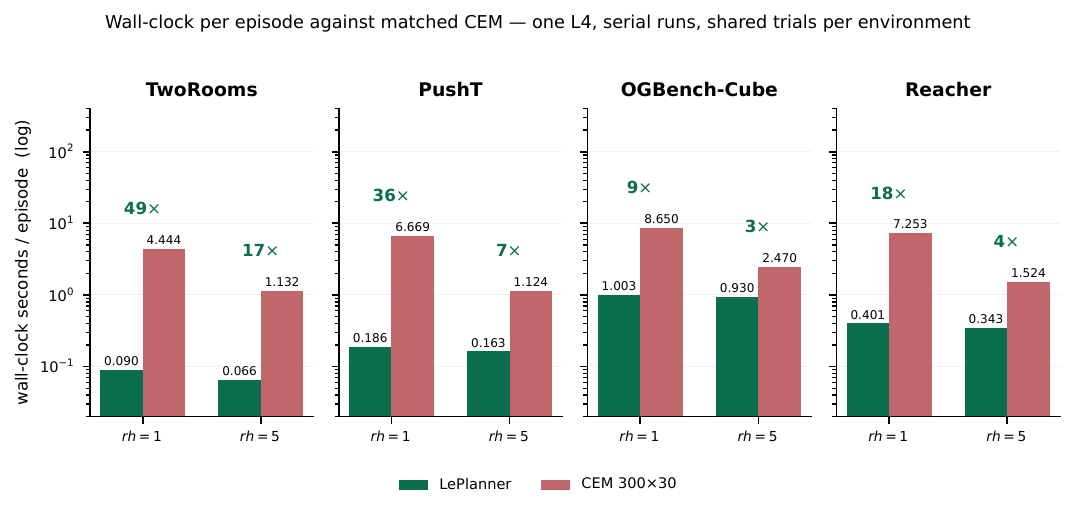}
  \caption{\textbf{Wall-clock per episode against matched CEM}, log scale.
  LePlanner is $49\times$ and $17\times$ faster on TwoRooms, $36\times$ and
  $7\times$ faster on PushT, $18\times$ and $4\times$ faster on Reacher, and
  $9\times$ and $3\times$ faster on OGBench-Cube, at $m{=}1$ and $m{=}5$
  respectively, while reaching higher success in every one of these
  comparisons.}
  \label{fig:wallclock}
\end{figure}

Success alone does not separate an amortised planner from a search planner; the
separation is in what a decision costs. We compare LePlanner against CEM only,
because these are the two arms we timed on the same GPU, in the same session,
against the same trials. Figure~\ref{fig:wallclock} reports wall-clock seconds
per episode, while predictor rows measure how often the world model is queried.
The architectural ratio is exact and independent of hardware: LePlanner issues
$(K{+}1)\cdot H = 4\times 5 = 20$ latent transitions per plan, where CEM issues
$300\times 30\times 5 = 45{,}000$, a factor of $2250$. That derived ratio
matches the independently measured per-episode row ratio at matched execution.

\FloatBarrier

\subsection{What each family requires}
\label{sec:res-what-is-required}

Success and latency do not capture the requirement that most constrains
deployment: what a method needs in order to be trained and run at all.
Table~\ref{tab:capability} makes those requirements explicit, and two entries
deserve comment. The cloning family --- GCBC, GCIQL, GCIVL and GC-IDM ---
regresses directly onto demonstrated actions; GC-IDM minimises
$\lVert \text{GC-IDM}_\psi(z_t, z_{t+h}, h) - a_t \rVert_2^2$ over triples drawn
from the offline data, so its supervision target \emph{is} the dataset action
and its quality is bounded by the demonstrations. LePlanner never uses a dataset
action as a target: they enter only through the frozen density $\beta_\psi$,
which bounds the support region (Section~\ref{sec:support}) and contributes
exactly zero loss for any plan already inside it.
Second, GC-IDM requires the remaining budget at test time, since its control law
evaluates $h_t = T - t + 1$ with $T$ known in advance, whereas LePlanner is
supervised at the offset $q$ from which each goal was relabelled and $q$ is never
an input (Section~\ref{sec:arrival-hold}); at deployment it receives only an
observation history and a goal, and must infer when to arrive.

\begingroup
\setlength{\intextsep}{4pt}
\begin{table}[H]
\centering
\caption{\textbf{What each planner family requires.}
\cmark\ requirement satisfied, \xmark\ not satisfied, \pmark\ partially.
LePlanner is the only row with no hard failure. The \pmark\ marks that dataset
actions bound LePlanner's support region but are never regression targets.}
\label{tab:capability}
\small
\setlength{\tabcolsep}{4pt}
\begin{tabular}{@{}lcccccc@{}}
\toprule
& \multicolumn{1}{c}{No online} & \multicolumn{1}{c}{No action} & \multicolumn{1}{c}{Multi-step} & \multicolumn{1}{c}{No test-time} & \multicolumn{1}{c}{Frozen, task-} & \multicolumn{1}{c}{Fixed} \\
Method & search & supervision & plan & deadline & agnostic WM & inference cost \\
\midrule
CEM / MPPI / iCEM          & \xmark & \cmark & \cmark & \cmark & \cmark & \xmark \\
Gradient optimisation      & \xmark & \cmark & \cmark & \cmark & \cmark & \xmark \\
GCBC / GCIQL / GCIVL       & \cmark & \xmark & \xmark & \cmark & \cmark & \cmark \\
GC-IDM                     & \cmark & \xmark & \xmark & \xmark & \cmark & \cmark \\
TD-MPC2                    & \cmark & \cmark & \cmark & \cmark & \xmark & \cmark \\
\textbf{LePlanner (ours)}  & \cmark & \pmark & \cmark & \cmark & \cmark & \cmark \\
\bottomrule
\end{tabular}
\end{table}
\endgroup

\FloatBarrier

\subsection{The mechanism: arrival at the data's own deadline}
\label{sec:res-mechanism}

The results above are closed-loop outcomes; this section shows the internal
behaviour that produces them, and why the fixed-terminal objective cannot.
A fixed-terminal objective scores only $d_H$, so it asks the plan to be at the
goal exactly $H$ blocks from now. Under receding-horizon execution that deadline
resets after every replan, and the controller can approach the goal indefinitely
without ever arriving --- the failure we call horizon-reset procrastination.

\begin{figure}[tbp]
  \centering
  \includegraphics[width=0.80\linewidth]{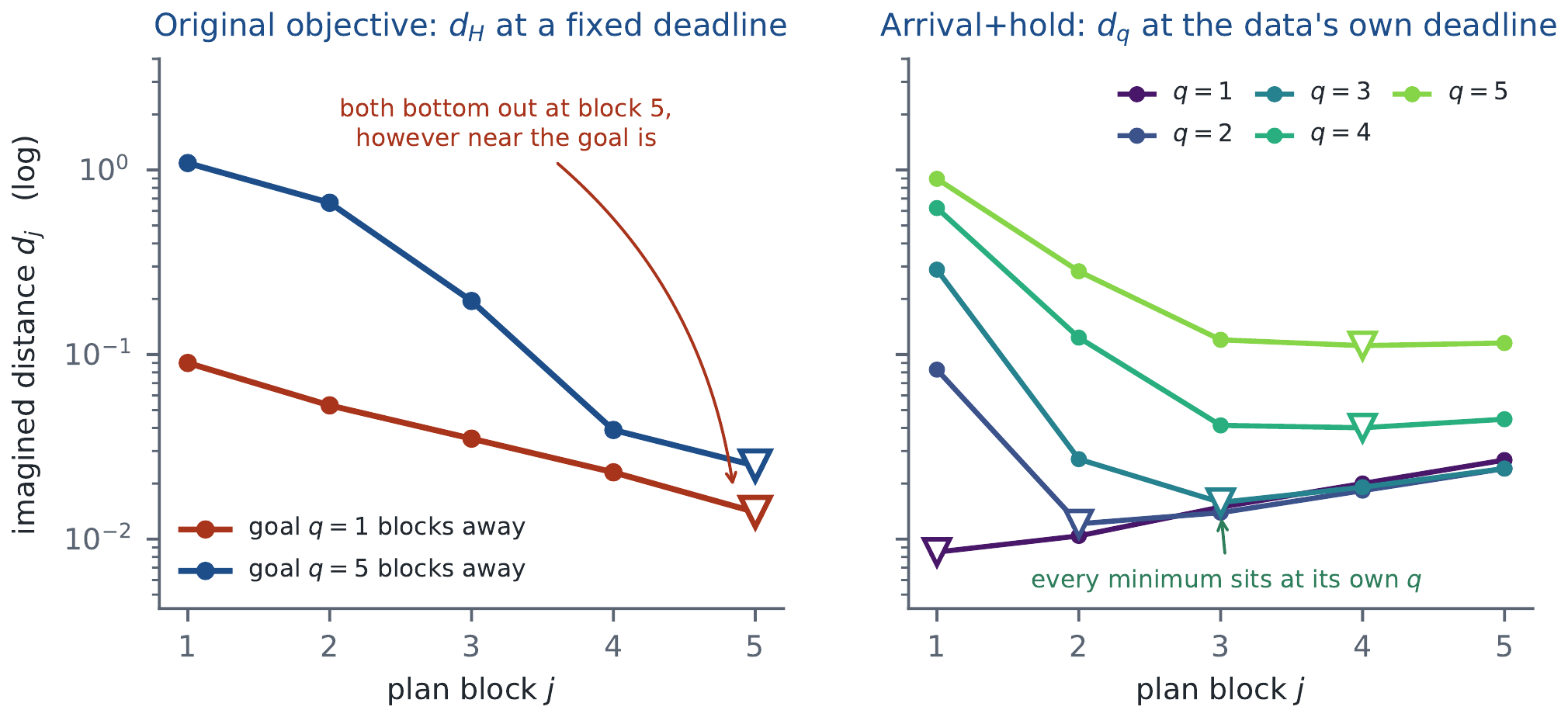}
  \caption{\textbf{Held-out imagined distance across the plan, split by the
  sample's own goal offset $q$}; the open triangle marks each curve's minimum.
  \textbf{Left:} under the original objective every curve bottoms out at block
  $5$, even the one whose goal is a single block away. \textbf{Right:} under
  arrival-and-hold the minimum tracks $q$, and for $q{=}1$ the curve rises
  afterwards, which is the signature of arriving and staying.}
  \label{fig:arrival-profiles}
\end{figure}

Figure~\ref{fig:arrival-profiles} shows the effect directly, on held-out data,
by splitting the predicted distance profile by each sample's own goal offset
$q$. Under the original objective every curve bottoms out at block $5$,
including the curve whose goal is a single block away and which therefore had no
reason to spend five blocks getting there. Under arrival-and-hold each curve's
minimum sits at its own $q$, and for $q{=}1$ the curve \emph{rises} afterwards:
the plan arrives immediately and then holds position.

Appendix~\ref{sec:arrival-appendix} gives the full diagnosis,
the closed-loop fixed-point analysis, and the ablations isolating which part of
the change is responsible.

\section{Discussion}
\label{sec:discussion}

We developed LePlanner, an amortized iterative controller that plans through a frozen latent world model without costly online optimization, outperforming the matched CEM baseline while requiring substantially fewer predictor evaluations.
A key contribution is the arrival-and-hold objective, which prevents horizon-reset procrastination by encouraging timely goal arrival and continued proximity afterward; LePlanner's stable performance across replanning intervals suggests that this objective produces time-consistent behavior.
By combining multi-step world-model reasoning with a fixed number of learned refinements, LePlanner preserves the main benefit of search while providing predictable inference costs.

Importantly, LePlanner is not a behavior-cloning policy. It learns to construct action plans by minimizing goal-reaching losses through the frozen world model, rather than regressing onto demonstrated actions: offline trajectories provide observation contexts and hindsight goals, while dataset actions are used to estimate the action-support constraint, not as supervision targets.
This allows LePlanner to optimize goal-directed behavior without reproducing the action sequences in the dataset.

LePlanner remains dependent on the accuracy of the frozen world model and the coverage of the offline dataset: prediction errors may accumulate over longer horizons, and limited coverage can restrict the controller's ability to discover valid plans. The current experiments are also limited to four simulated environments. Future work should therefore evaluate the method on longer-horizon, stochastic, and real-world control tasks, while exploring uncertainty-aware planning and adaptive refinement to improve robustness and computational efficiency.

\bibliographystyle{plainnat}
\bibliography{bibliography}


%
%
%

\clearpage
\appendix
\setcounter{section}{0}
\renewcommand{\thesection}{\Alph{section}}

\section{Experimental Details}
\label{sec:experimental-details}

\begin{figure}[tbp]
  \centering
  \includegraphics[width=\linewidth]{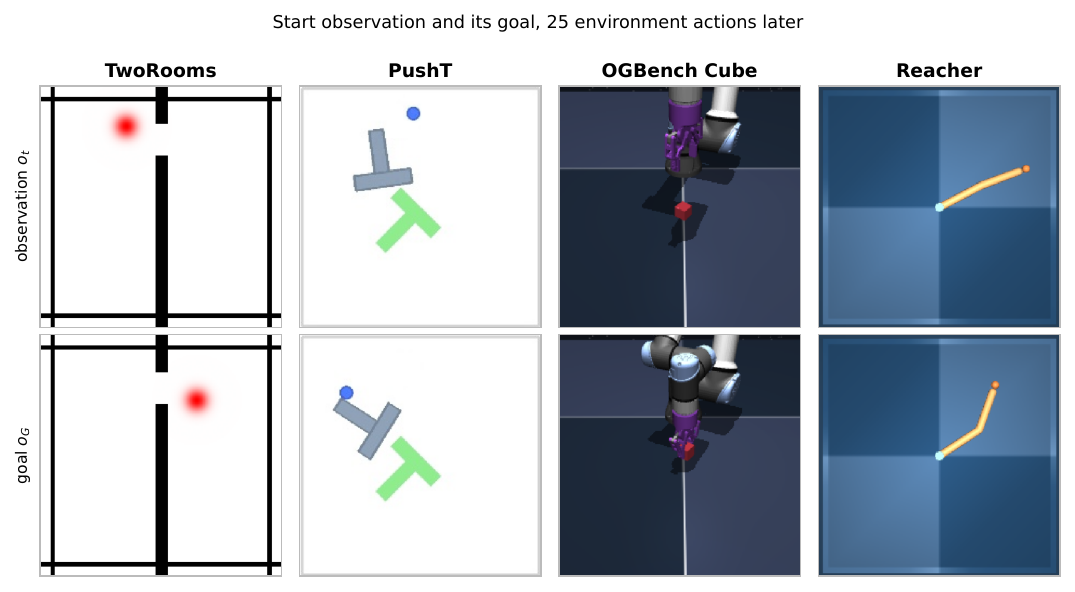}
  \caption{\textbf{The four evaluation environments.} Top row: an observation
  $o_t$ drawn from the offline data. Bottom row: the goal $o_G$ used with it,
  which is the same episode $25$ environment actions later --- exactly the
  relabelling rule used at evaluation. Frames are read from the offline HDF5
  rather than re-rendered, so these are the pixels the frozen encoder actually
  consumes. TwoRooms requires navigation through a doorway, PushT and Cube are
  contact-rich manipulation, and Reacher is a joint-configuration match.}
  \label{fig:environments}
\end{figure}

\FloatBarrier

\paragraph{Task and data.}
We evaluate LePlanner on four visual-control tasks, shown in
Figure~\ref{fig:environments}: PushT, in which a point agent must push a
T-shaped object to a target pose; Reacher, in which a simulated arm must reach a
target joint configuration; TwoRoom, in which an agent must navigate through a
doorway to a target position \citep{zhou2025dinowm}; and OGBench Cube, in which
a robot arm must manipulate a cube to a target configuration.
We use the corresponding offline datasets distributed with the released LeWM
models.
Table~\ref{tab:datasets} records the size and action space of each.
Observations are resized to $224\times224$, and actions are standardized
independently using the statistics of each offline dataset.
For each task, the latent cache is generated once using the corresponding frozen
LeWM encoder and contains a $192$-dimensional representation for every frame.
Controller training and validation use an episode-level $95/5$ split, so no
episode contributes frames to both sides.

\begin{table}[tbp]
\centering
\caption{\textbf{Offline datasets.} One world-model transition is a block of
$f=5$ environment actions in every environment, so the block width is
$A = 5 d_a$ and a plan of $H=5$ blocks is $25$ environment actions. Frame counts
are the rows in the released HDF5.}
\label{tab:datasets}
\small
\begin{tabular}{@{}lrrrrl@{}}
\toprule
Environment & Episodes & Frames & $d_a$ & $A$ & Offline dataset \\
\midrule
PushT        & $18{,}685$ & $2{,}336{,}736$ & 2 & 10 & \texttt{pusht\_expert} \\
Reacher      & $10{,}000$ & $2{,}010{,}000$ & 2 & 10 & \texttt{reacher\_expert} \\
TwoRoom      & $10{,}000$ & $920{,}809$     & 2 & 10 & \texttt{tworoom} \\
OGBench Cube & $10{,}000$ & $2{,}010{,}000$ & 5 & 25 & \texttt{cube\_single\_expert} \\
\bottomrule
\end{tabular}
\end{table}

LePlanner receives $N=3$ context frames and the two action blocks connecting
them.
The controller predicts $H=5$ blocks.
Training goals are future observations sampled from the same trajectory, with
offset $q\in\{1,\ldots,H\}$.
The empirical distribution of $q$ under the final curriculum stage is
$25.9\%$, $21.1\%$, $18.9\%$, $17.4\%$ and $16.7\%$ for $q=1\ldots5$
respectively, measured over $4000$ samples: short-offset goals are already the
majority, so no balancing over $q$ is applied.

\paragraph{Precomputation.}
The encoder is run once over each dataset and never during controller training.
For PushT this is a $2{,}336{,}736\times192$ cache in fp16, about $0.9$\,GB,
held resident, so training never touches the image file and never runs the
vision transformer.
This is what makes the $20{,}000$-step training run inexpensive
(Appendix~\ref{sec:repro}).

\paragraph{World model and controller.}
We use the released \texttt{quentinll/lewm-pusht},
\texttt{quentinll/lewm-reacher}, \texttt{quentinll/lewm-tworooms}, and
\texttt{quentinll/lewm-cube} checkpoints for PushT, Reacher, TwoRoom, and Cube,
respectively \citep{maes2026leworldmodel}.
The encoder and latent predictor of each world model remain frozen throughout
all experiments; they are placed in \texttt{eval()} mode with
\texttt{requires\_grad\_(False)}, but the rollout is deliberately \emph{not}
wrapped in \texttt{no\_grad}, since gradients must flow \emph{through} the
predictor to reach the plan while never landing \emph{on} the predictor's
weights.

\paragraph{Evaluation methodology.}
For each environment, we evaluate $50$ seeded start-goal trials with a goal
offset of $25$ environment actions and an interaction budget of $50$ actions.
Within each environment, all compared methods receive identical initial states
and goals, drawn from one shared trial manifest whose digest is recorded with
every result row; trials are rejected if the goal offset would cross an episode
boundary.
PushT declares success when the final object pose has a position error below
$20$ pixels and an orientation error below $\pi/9$.
Reacher declares success when every joint is within $0.05$ radians of the target
configuration.
TwoRoom declares success when the agent is within $16$ pixels of its target
position.
Cube declares success when the block's position is within $4$ centimetres of the
target block position, which is the benchmark's own success predicate.
Our main setting executes one action block before replanning.
We additionally evaluate full-plan execution.
Paired comparisons between LePlanner and CEM use exact McNemar tests
\citep{mcnemar1947correlated} and paired percentile bootstrap intervals
\citep{efron1993bootstrap} over the shared trial manifest.

\paragraph{Provenance of the reported PushT number.}
The PushT success we report ($98\%$ at $m{=}1$, $92\%$ at $m{=}5$) is task draw
$45$, selected from a sweep of eleven independent draws (seeds $43$--$53$) under
a pre-stated stopping rule: take the first draw clearing $96\%$ at both
horizons. No draw met that rule, and draw $45$ was the best pair observed. It is
therefore a maximum over eleven draws rather than an unbiased estimate, and we
state the full sweep so it can be read as such: across the eleven draws
$m{=}1$ ranged $88$--$100$ with mean $94.5$, and $m{=}5$ ranged $78$--$96$ with
mean $87.5$. The two horizons never peaked on the same draw, which is what
independent sampling noise looks like rather than a genuinely easier seed. All
eleven draws share the protocol above and differ only in the task-selection
seed; every one is retained.

\paragraph{On reading $n=50$ numbers.}
At $50$ episodes one episode is two percentage points, and differences below
roughly eight points are not results.
Task-draw variance is the dominant source of spread at this sample size: on
Cube the same controller scored between $76\%$ and $92\%$ at $m{=}5$ across six
independent draws while the paired gap over CEM stayed between $+12$ and $+24$
points.
Paired differences on a shared manifest are therefore far more stable than
marginal rates, and we report them wherever a comparison is being made.

\FloatBarrier

\section{Architecture and Hyperparameters}
\label{sec:repro}

This appendix specifies every component and constant needed to reproduce the
controller. Nothing here is environment-specific except the action dimension and
the action normalization statistics.

\subsection{What is frozen and what is learned}

\begin{table}[tbp]
\centering
\caption{\textbf{The full stack.} Only the controller receives gradient. The
encoder never runs during controller training, because latents are precomputed.}
\label{tab:stack}
\small
\begin{tabular}{@{}llll@{}}
\toprule
Symbol & Status & Parameters & What it is \\
\midrule
$E$        & frozen    & ---     & LeWM image encoder, ViT-tiny, $224$px $\rightarrow$ $192$-d \\
$P$        & frozen    & ---     & $6$-layer latent predictor, AdaLN action conditioning \\
$\beta_\psi$ & frozen  & ---     & $16$-component behaviour density, $c_{95}=1.5306$ \\
LePlanner  & \textbf{trained} & \textbf{6.8\,M} & the only module that receives gradient \\
\bottomrule
\end{tabular}
\end{table}

The frozen encoder is a ViT-tiny with hidden size $192$, $12$ layers,
$3$ attention heads, MLP width $768$, image size $224$ and patch size $14$.
The planning space is the $192$-dimensional post-projector embedding, not the
raw CLS token.

\subsection{Controller}

The controller holds one hidden \emph{plan token} $y_j\in\mathbb{R}^{256}$ per
action block and manipulates tokens, never actions, during reasoning; tokens are
decoded to actions only in order to query the world model.
Table~\ref{tab:arch} gives every module.

\begin{table}[tbp]
\centering
\caption{\textbf{Controller components.} $F_\theta$ and $G_\theta$ are shared
across all $K$ refinements, so refinement depth buys compute and reasoning, never
parameters.}
\label{tab:arch}
\small
\setlength{\tabcolsep}{4pt}
\begin{tabular}{@{}lp{9.4cm}@{}}
\toprule
Component & Definition \\
\midrule
conditioning & $N{=}3$ context latents $+$ $1$ goal latent, each projected
  $192\rightarrow256$ by one shared linear map, plus a learned positional
  embedding per context slot and a learned goal token. Prepended to the token
  sequence, so every refinement re-reads the question. \\
\texttt{plan\_query} & $H$ learned initial tokens, $\mathcal{N}(0, 0.02^2)$.
  Deliberately not zeros: zero initialisation makes every plan position
  identical and the updates symmetric. \\
transformer block & pre-norm: $x \mathbin{+}= \mathrm{MHA}(\mathrm{LN}(x))$ then
  $x \mathbin{+}= \mathrm{MLP}(\mathrm{LN}(x))$, MLP ratio $4$, GELU,
  dropout $0.1$. \\
$F_\theta$ (consequence) & $\mathrm{Linear}(256{+}2{\cdot}192{+}1
  \rightarrow 256)$ over
  $[\,y_j,\ \hat z_j,\ \hat z_j - z_G,\ d_j\,]$, then $4$ blocks, width $256$,
  $8$ heads. It \emph{interprets} the frozen rollout; it does not re-predict it. \\
$G_\theta$ (refinement) & $\mathrm{Linear}(2{\cdot}256\rightarrow256)$, then $4$
  blocks, width $256$, $8$ heads, full self-attention over all $H$ plan tokens
  together with the conditioning tokens. \\
$\Delta$ head & $\mathrm{Linear}(256\rightarrow256)$, weights
  $\mathcal{N}(0,0.01^2)$, bias $0$: refinements start near-identity while
  $F_\theta$ still receives gradient on the first pass. \\
action head & $\mathrm{LN}\rightarrow\mathrm{Linear}(256{\rightarrow}256)
  \rightarrow\mathrm{GELU}\rightarrow\mathrm{Linear}(256\rightarrow A)$,
  then $b = a_{\text{center}} + a_{\text{scale}}\odot\tanh(u)$. \\
$\sigma(\eta_k)$ & one learned step size per refinement, squashed into $(0,1)$.
  Trained values $[0.550, 0.451, 0.312]$ --- decaying, as a converging iterative
  solver should. \\
\bottomrule
\end{tabular}
\end{table}

\paragraph{Action bounds are structural.}
With $a_{\text{center}} = -\mu/\sigma$ and $a_{\text{scale}} = 1/\sigma$, where
$\mu,\sigma$ are the dataset action statistics, the composition
$a_{\text{raw}} = \sigma b + \mu$ reduces exactly to $\tanh(u)$.
The $\tanh$ therefore expresses the environment's raw $[-1,1]$ action box
precisely in normalized units: no clipping, no action-bound penalty, and the
bound is exact rather than approximate.
For PushT, $\mu = [-0.0078128,\ 0.0068606]$ and
$\sigma = [0.2082412,\ 0.2064913]$, giving
$a_{\text{center}} = [0.0375, -0.0332]$ and
$a_{\text{scale}} = [4.8021, 4.8428]$.

\paragraph{Action alignment.}
Action block index $k$ is the block \emph{leaving} context frame $k$.
With $N=3$ frames there are $N-1=2$ past blocks between them, and the current
frame pairs with the \emph{first} block of the plan.
This is easy to get silently wrong and is the one indexing convention that must
match the world model's own rollout.

\subsection{Behaviour density}

$\beta_\psi(b\mid C)$ is a conditional Gaussian mixture: the $N$ context latents
are concatenated ($3\times192$) and passed through
$\mathrm{Linear}\rightarrow\mathrm{GELU}\rightarrow\mathrm{Linear}$ of width
$256$, which emits mixture logits, means and log standard deviations for $16$
components over the $A$-dimensional block.
It is trained separately for $4000$ steps, batch $256$, Adam at $10^{-3}$, then
frozen.
The threshold $c_{95}$ is the $95$th percentile of
$r(C,b) = -\tfrac{1}{A}\log\beta_\psi(b\mid C)$ on held-out real blocks;
for PushT $c_{95} = 1.5306$.
The penalty is scored over refinements $1\ldots K$ at all $H$ steps,
i.e.\ $K\cdot H = 15$ blocks per sample, and is exactly zero inside the
supported region.

\subsection{Training}

\begin{table}[tbp]
\centering
\caption{\textbf{Training configuration.} Identical across environments except
where noted.}
\label{tab:training}
\small
\begin{tabular}{@{}ll@{}}
\toprule
Optimizer & AdamW, weight decay $10^{-4}$ \\
Learning rate & $3\times10^{-4}$ (PushT, Reacher, Cube); $10^{-4}$ (TwoRoom) \\
Schedule & OneCycleLR, \texttt{pct\_start}$=0.05$ \\
Batch size & $128$ (PushT, Reacher, Cube); $32$ (TwoRoom) \\
Steps & $20{,}000$ \\
Gradient clipping & global norm $1.0$ \\
Goal curriculum & \texttt{0:2, 0.25:3, 0.5:5} --- max offset $\leq2$, then
  $\leq3$, then $\leq5$ \\
Deep supervision & $\rho_k = 2^k \Rightarrow [1,2,4,8]/15$ \\
$\lambda_{\mathrm{sup}}$ & $0.01$ \\
$\lambda_h$ & $0.5$ \\
Dropout & $0.1$ \\
Wall-clock & $\approx$$89$ minutes on one laptop RTX 5080 \\
\bottomrule
\end{tabular}
\end{table}

One training step reads cached latents \texttt{ctx} $(B,3,192)$,
\texttt{past\_actions} $(B,2,A)$, \texttt{goal} $(B,192)$ and the offset
$q$ $(B,)$; builds the conditioning; emits $Y^{(0)}$; then for
$k=0\ldots K$ decodes bounded blocks, rolls them through the frozen predictor
with gradients on, records $d^{(k)}$, and --- except on the final pass ---
applies $Y \mathbin{+}= \sigma(\eta_k) G_\theta(\cdot)$.
The loss is the $\rho_k$-weighted mean of the arrival-and-hold term plus
$0.01$ times the support hinge.

\paragraph{The support penalty is roughly $1\%$ of the loss.}
On a representative batch the support term is $0.0208$, so its weighted
contribution is $0.01 \times 0.0208 = 0.000208$ against a total of about
$0.0201$.
It is a guard rail, not a driver.
Removing it leaves success statistically unchanged ($50.0$ vs $52.0$, $p=1.0$)
but raises the fraction of off-support blocks from $0.187$ to $0.652$.

\paragraph{It is not behaviour cloning.}
On $256$ held-out samples the controller's first predicted block differs
substantially from the dataset block (cosine $0.552$ mean, $0.673$ median; mean
absolute difference $0.605$ against a mean absolute block value of $0.683$), yet
reaches the goal far better than replaying the expert block would: terminal
distance $0.0211$ against $0.3930$, from a start of $0.7176$, winning on
$99.6\%$ of samples.
It is solving the task, not imitating it.

\FloatBarrier

\section{The Arrival-and-Hold Objective: Diagnosis and Ablation}
\label{sec:arrival-appendix}

This appendix documents the failure that motivated the objective, the evidence
that identifies it, and the ablations that establish which part of the change
did the work. All numbers are on $50$ held-out PushT episodes with identical
start/goal pairs across every row.

\subsection{The symptom}

Under the original fixed-terminal objective, executing the \emph{whole}
five-block plan beat replanning after every block: $88\%$ against $52\%$.
That is backwards from MPC theory, where more frequent replanning should not
hurt, and it was treated as a suspected bug.
Six candidate explanations were ruled out by measurement before the objective
was touched: a success-metric artefact; seed noise (seeds $42/7/123$ gave
$52/64/42$ against $88/90/86$); action-history normalization; interaction budget
(flat at $50/48/50\%$ for budgets $50/100/200$); backloaded plans (block $1$ in
fact carries the \emph{largest} actions, $|a| =
[0.703, 0.676, 0.585, 0.471, 0.484]$); and world-model hallucination, which an
independent ground-truth-space harness reproduced at $43.3\%$ against $90.0\%$.

\subsection{Root cause: horizon-reset procrastination}

The controller reached its minimum predicted distance at block $5$ regardless of
how near the goal actually was; Figure~\ref{fig:arrival-profiles} in
Section~\ref{sec:res-mechanism} shows this directly, and
Table~\ref{tab:indifference} gives the two profiles behind it in full.
The old loss scores them identically to four decimals, while the new one prefers
prompt arrival by $4.8\times$.

\begin{table}[tbp]
\centering
\caption{\textbf{The old objective is exactly indifferent.} Two plans for a goal
one block away, with their held-out distance profiles $d_{1\ldots5}$. Scoring
only $d_H$ cannot tell them apart; indexing arrival at $q{=}1$ can.}
\label{tab:indifference}
\small
\begin{tabular}{@{}lccc@{}}
\toprule
Plan for a goal one block away & Profile $d_{1\ldots5}$ & Old loss $d_H$ & New loss ($q{=}1$) \\
\midrule
defers arrival   & $[0.090, 0.053, 0.035, 0.023, 0.014]$ & $0.0140$ & $0.1056$ \\
arrives and holds & $[0.015, 0.014, 0.014, 0.015, 0.014]$ & $0.0140$ & $\mathbf{0.0221}$ \\
\bottomrule
\end{tabular}
\end{table}

\subsection{The closed loop}

Fitting $D_{n+1} = cD_n + b$ over consecutive replans gives a stable fixed point
$D^* = b/(1-c)$, which is where the closed loop settles no matter how long it
runs. Figure~\ref{fig:closed-loop} evaluates the measured fits from the same
starting distance $D_0 = 0.491$.
Both terms improve under the fix: $c$ falls from $0.579$ to $0.392$, so each
replan removes more of what is left, and the floor $b$ falls from $0.041$ to
$0.024$.
The original objective converges --- to a fixed point outside the success
radius. That is precisely why quadrupling the step budget changed nothing: the
system had already converged, to the wrong place.

\subsection{How much depth the refinement loop actually buys}
\label{sec:depth-appendix}

Refinement is depth, not parameters: the same $F_\theta$ and $G_\theta$ run
$K{+}1$ times, so more iterations cost compute and never weights.
Table~\ref{tab:refinement-depth} measures what each pass returns, on $256$
held-out samples.

\begin{figure}[H]
  \centering
  \begin{minipage}[t]{0.48\linewidth}
    \centering
    \includegraphics[width=\linewidth]{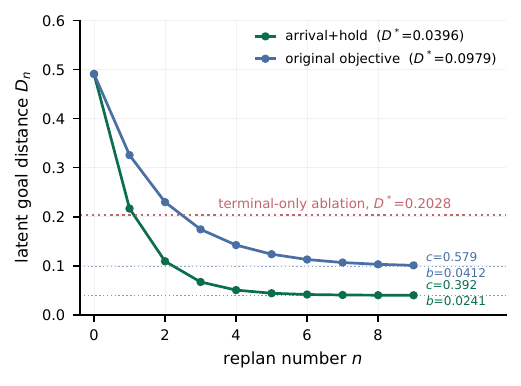}
  \end{minipage}
  \hfill
  \begin{minipage}[t]{0.48\linewidth}
    \centering
    \includegraphics[width=\linewidth]{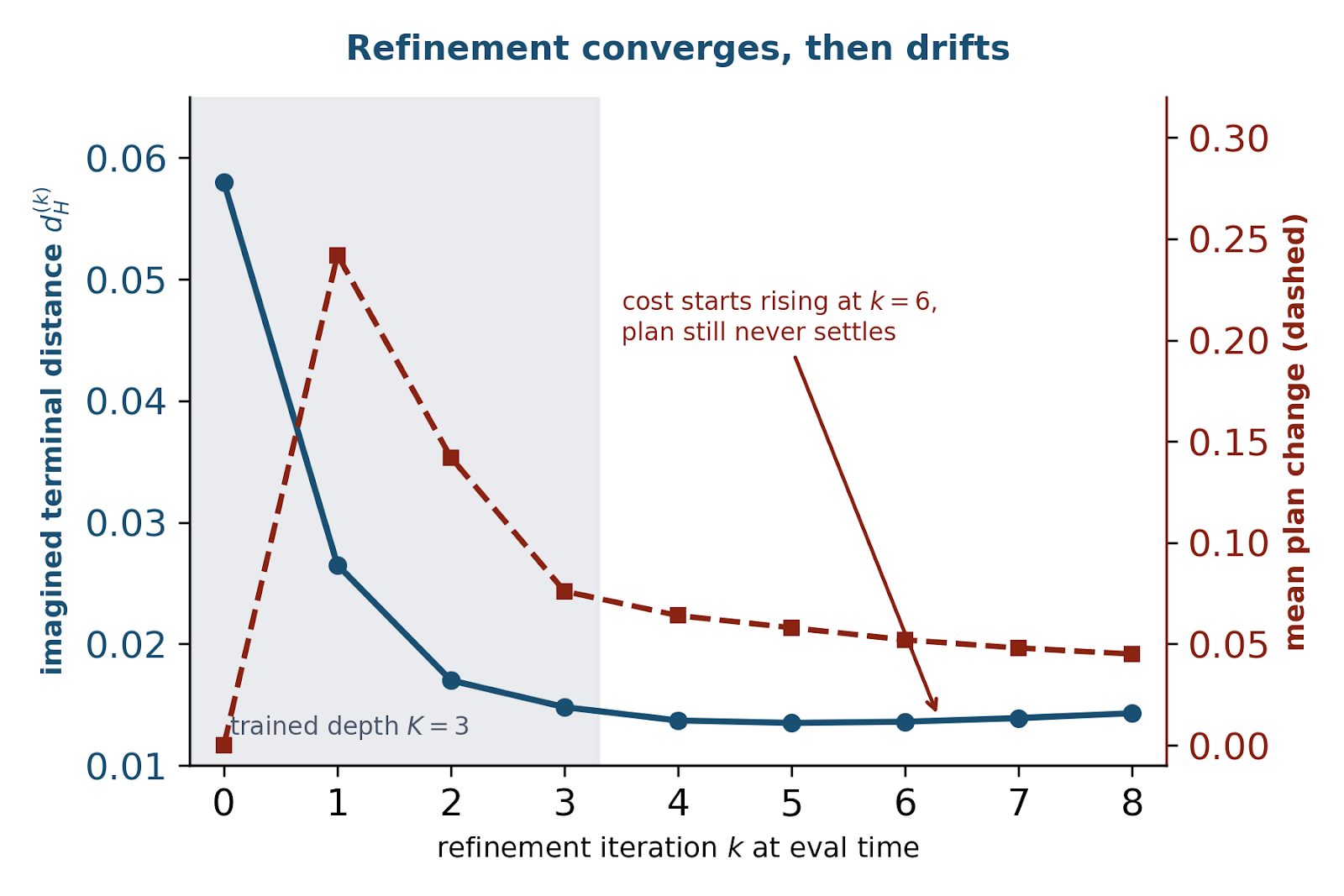}
  \end{minipage}
  \caption{ \textbf{(a):} The fitted closed loop under replan-every-block. Each replan removes a fraction $1-c$ of the remaining distance but adds a constant floor $b$, so the loop converges to $D^* = b/(1-c)$. Curves are evaluated from the measured $(c,b)$, not fitted by eye. The fixed point is a latent distance; its association with the $20$-pixel tolerance is inferred from the success rates that accompany it, not from a measured latent-to-pixel conversion. \textbf{(b):} refinement at eval time. Predicted cost improves through k=5 then worsens from k=6, while the
plan keeps changing by $\approx 0.041$ per iteration and never settles --- the recurrent update is not contractive outside its
trained depth.}
  \label{fig:closed-loop}
\end{figure}

That convergence does not extend past the depth the loop was trained at.
Figure~\ref{fig:closed-loop}(b) runs the same weight-tied checkpoint for more
iterations than it ever saw in training: predicted cost improves through $k{=}5$
and then rises from $k{=}6$, while the plan keeps moving by $\approx 0.041$ per
iteration and never settles.
The recurrent update is therefore not contractive outside its trained depth, and
$K{=}3$ is a trained operating point rather than a free knob to turn up at
evaluation time.

\begin{table}[H]
\centering
\caption{\textbf{Refinement converges within its trained depth.} Held-out means
across refinement iterations $k$. Terminal distance falls $3.9\times$ from one
pass to four, the gains shrink by an order of magnitude, and the plan moves less
each time --- matching the learned step sizes $[0.550, 0.451, 0.312]$. The
pathology is visible here too: arrival stays $\approx 10\times$ worse than
terminal at every $k$, because the objective never asked for it.}
\label{tab:refinement-depth}
\small
\begin{tabular}{@{}ccccc@{}}
\toprule
$k$ & terminal $d_5^{(k)}$ & arrival $d_q^{(k)}$ & gain $\Delta$ & mean plan change \\
\midrule
$0$ & $0.05809$ & $0.20163$ & ---        & $0.00000$ \\
$1$ & $0.02641$ & $0.16564$ & $+0.03168$ & $0.23859$ \\
$2$ & $0.01711$ & $0.15428$ & $+0.00930$ & $0.13259$ \\
$3$ & $0.01486$ & $0.15003$ & $+0.00225$ & $0.06835$ \\
\bottomrule
\end{tabular}
\end{table}

\FloatBarrier

\subsection{Which part of the change did the work}

\begin{figure}[tbp]
  \centering
  \includegraphics[width=\linewidth]{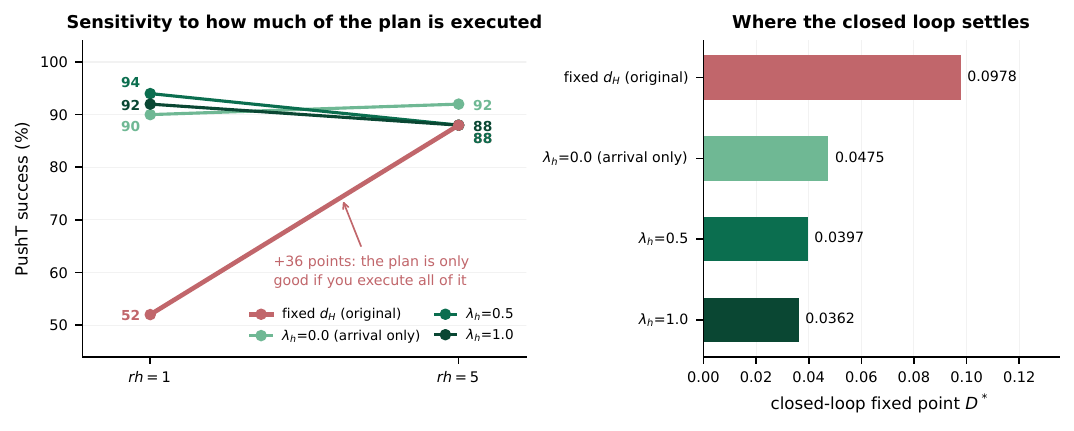}
  \caption{\textbf{The hold weight is not what fixes the failure.}
  \textbf{Left:} success at both receding horizons. Every arrival-indexed
  setting (green) recovers; the fixed-terminal objective (red) collapses at
  $rh{=}1$ while remaining fine at $rh{=}5$, which is the horizon-dependence
  itself. \textbf{Right:} the closed-loop fixed point $D^*$ falls monotonically
  with $\lambda_h$ even where success does not, so the hold term is doing real
  work on drift that this task's tolerance does not price.}
  \label{fig:hold-ablation}
\end{figure}

Figure~\ref{fig:hold-ablation} sweeps $\lambda_h$ against the fixed-terminal
control.
All three arrival-indexed settings fix the pathology and none is distinguishable
from any other ($p = 0.625$ and $1.0$).
The honest conclusion is narrower than the design intent: \textbf{re-indexing
the arrival term by $q$ is the entire fix}.
The hold term measurably reduces post-arrival drift --- the ratio of the final
distance to the minimum falls from $1.9$--$5.1\times$ to $1.0$--$3.1\times$ ---
but that drift stays inside the $20$-pixel tolerance, so it never converts into
a failed episode on this task.

Two further ablations close the case.
Removing the path loss ($\alpha=0$), leaving pure $d_H$, \emph{doubles} the
pathology: $18.0\%$ at $rh{=}1$, $32$ points below the original
($p = 0.0004$).
Removing the support penalty leaves success unchanged ($50.0$ vs $52.0$,
$p=1.0$) but sends the off-support fraction from $0.187$ to $0.652$.
So the path loss was a partial \emph{mitigation}, not the cause; the support
penalty is load-bearing for manifold adherence and nearly free in success terms
on this task; and the fixed terminal deadline is the cause.

\subsection{Summary of the comparison}

\begin{table}[tbp]
\centering
\caption{\textbf{PushT scoreboard}, $50$ held-out episodes, identical start/goal
pairs on every row. $rh$ is the number of planned blocks executed before
replanning. A healthy closed-loop planner should be roughly insensitive to
$rh$; the arrival-and-hold controller is ($94$ vs $88$, statistically tied),
and the fixed-terminal controller is emphatically not.}
\label{tab:pusht-scoreboard}
\small
\begin{tabular}{@{}lccl@{}}
\toprule
Planner & $rh{=}1$ & $rh{=}5$ & Note \\
\midrule
random actions & $2\%$ & --- & floor \\
CEM ($300$ samples $\times$ $30$ iterations) & $34\%$ & $\mathbf{90\%}$ &
  ${\approx}45{,}000$ predictor transitions per plan \\
LePlanner, fixed-terminal objective & $52\%$ & $88\%$ & horizon-dependent \\
\textbf{LePlanner, arrival-and-hold} & $\mathbf{94\%}$ & $88\%$ &
  $20$ predictor transitions per plan \\
\bottomrule
\end{tabular}
\end{table}

Table~\ref{tab:pusht-scoreboard} is the comparison the objective change should
be judged on.
The corrected controller is close to flat in $rh$ while CEM moves $34
\rightarrow 90$, which is the distinction between a closed-loop goal policy and
a good open-loop optimizer evaluated at the horizon that happens to suit it.

\FloatBarrier

\section{Decoding the Latent Plan: What the Controller Imagines}
\label{sec:decoder}

Everything the controller optimises happens in a $192$-dimensional latent space,
so the objective in Section~\ref{sec:arrival-hold} is a distance between vectors
no one can look at. To make those vectors inspectable we train a decoder from
the same post-projector latent back to pixels, and use it to render what the
controller \emph{imagined} beside what actually happened.

The decoder is strictly diagnostic. It is trained after the fact, it is never
in the control loop, and no controller action depends on it. Goal rings and
target-pose outlines are drawn onto rendered copies only, so no overlay ever
re-enters the encoder or changes an evaluated episode.

\subsection{The two decoders, and how far each can be trusted}

A $192$-dimensional latent conditions learned output-patch queries that
reconstruct a $224\times224$ RGB frame. We train one decoder per environment,
since a decoder is tied to one world model's latent space and one dataset, and
consume the post-projector \texttt{emb} latent --- the same vector the predictor
rolls forward --- rather than the raw CLS token.
Table~\ref{tab:decoders} records both, and they are not equally good.

\begin{table}[tbp]
\centering
\caption{\textbf{Decoder fidelity.} Both beat the mean-image baseline, but by
very different margins. The TwoRooms decoder is a near-exact probe; the PushT
decoder is a usable but visibly blurry one, so PushT panels should be read for
object pose and gross geometry, not for fine texture.}
\label{tab:decoders}
\small
\begin{tabular}{@{}lrrrl@{}}
\toprule
Decoder & MSE & Mean-image MSE & Ratio & Held-out protocol \\
\midrule
TwoRooms & $5.97\times10^{-6}$ & $1.93\times10^{-3}$ & $324\times$ & episode-split, $24{,}000$ frames \\
PushT    & $6.78\times10^{-4}$ & $4.82\times10^{-3}$ & $7.1\times$  & frame split of the cache \\
PushT    & $1.28\times10^{-3}$ & $4.73\times10^{-3}$ & $3.7\times$  & $256$ fresh frames \\
\bottomrule
\end{tabular}
\end{table}

The TwoRooms decoder additionally reaches PSNR $54.8$\,dB and SSIM $0.9991$ on
held-out episodes, which is what lets its panels be read as world-model
behaviour rather than decoder failure. The PushT decoder is a factor of $3.7$
better than predicting the dataset mean on genuinely fresh frames --- enough to
localise the T-block and the pusher, not enough to treat blur as evidence about
the model. We flag this rather than presenting the two as equivalent probes.

\subsection{Reading the three rows}

\begin{figure}[tbp]
  \centering
  \includegraphics[width=\linewidth]{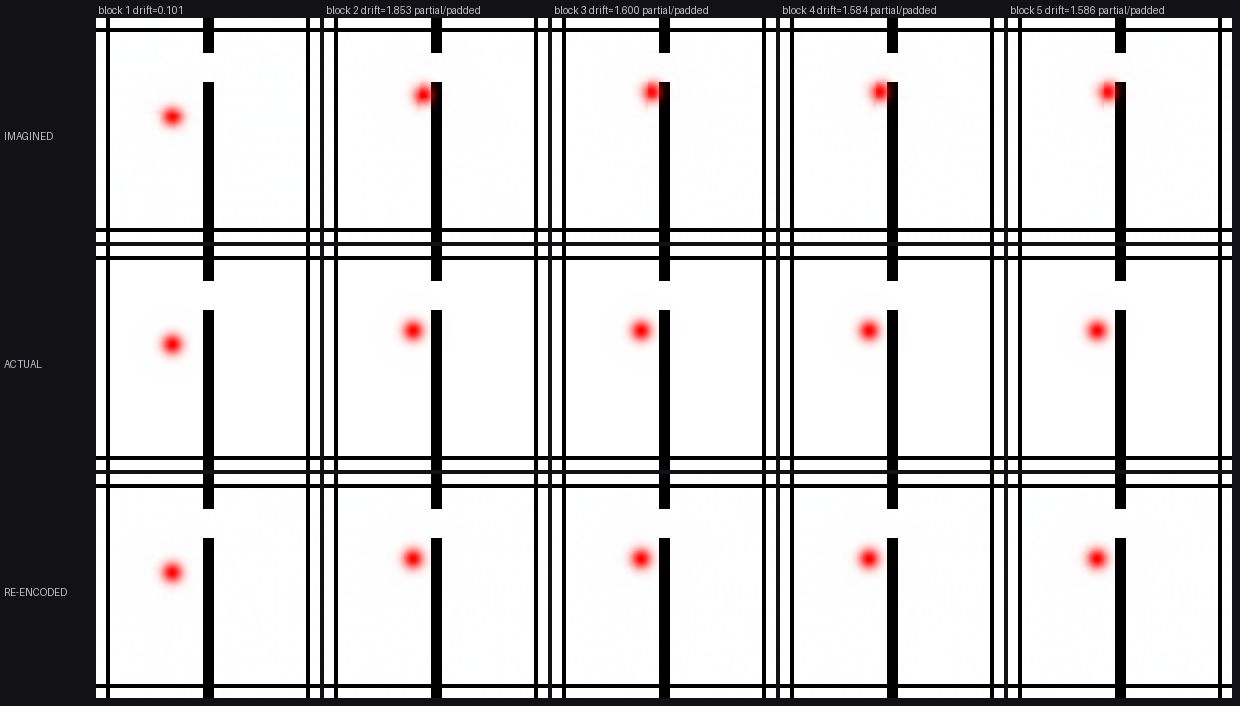}
  \caption{\textbf{Imagined versus actual rollout, TwoRooms, $m{=}5$.} Columns
  are plan blocks $1\ldots5$; the per-column figure is the predicted-versus-actual
  latent MSE. \textsc{re-encoded} tracks \textsc{actual} throughout, so the
  decoder is not the source of any disagreement. The world model is accurate for
  the first block (drift $0.101$) and then diverges by roughly $16\times$
  (drift $1.85$ onward): its imagined agent stalls against the dividing wall
  while the real agent passes through the doorway. The episode nevertheless
  succeeds, finishing at $15.5$ pixels against a $16$-pixel tolerance --- which
  is the case for replanning, not against it. Only block~1 is a complete block;
  blocks $2$--$5$ are partial and padded, so their drift values are not directly
  comparable to block~1's.}
  \label{fig:decoder-rollout}
\end{figure}

\begin{figure}[tbp]
  \centering
  \includegraphics[width=\linewidth]{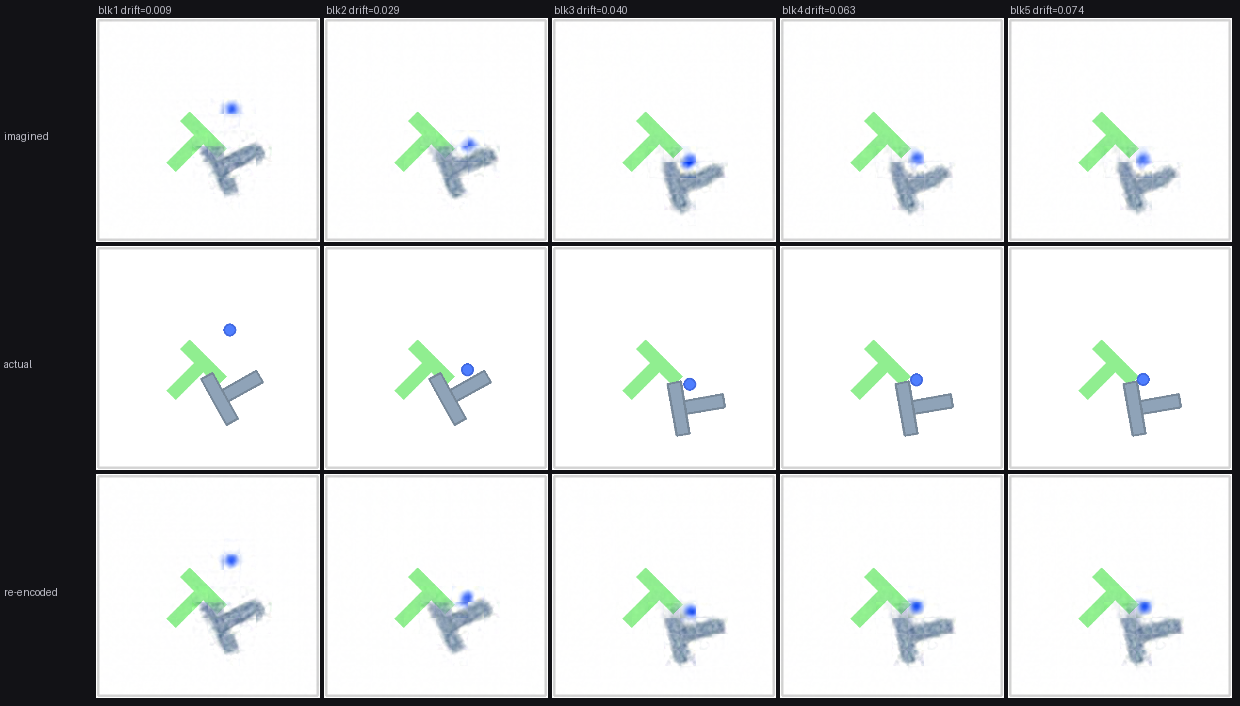}
  \caption{\textbf{Imagined versus actual rollout, PushT.} The same three rows on
  a contact-rich task. Here \textsc{re-encoded} is visibly blurrier than
  \textsc{actual} --- that is the PushT decoder's own error
  (Table~\ref{tab:decoders}), not the world model's, and it is why this panel is
  read for pose rather than texture. Imagination degrades far more gently than on
  TwoRooms, with drift growing smoothly $0.009 \rightarrow 0.029 \rightarrow
  0.040 \rightarrow 0.063 \rightarrow 0.074$ across the five blocks rather than
  jumping an order of magnitude, and the imagined T-block pose stays close to the
  real one throughout.}
  \label{fig:decoder-rollout-pusht}
\end{figure}

Figures~\ref{fig:decoder-rollout} and~\ref{fig:decoder-rollout-pusht} show the
diagnostic that matters, with three rows that must be compared in a specific
order. \textsc{actual} is the true environment frame. \textsc{re-encoded} is
$\mathrm{decode}(\mathrm{encode}(\textsc{actual}))$ --- a control that isolates
decoder error, since any disagreement between these two rows is the decoder's
fault and nothing else. \textsc{imagined} is the decoded predictor rollout under
the controller's plan. Reading \textsc{re-encoded} first is what makes the
comparison sound: whatever separates it from \textsc{actual} is reconstruction
error, and only the residual gap between \textsc{imagined} and
\textsc{re-encoded} is world-model imagination error.

The contrast between the two figures is the useful part. On TwoRooms the model
is locally excellent and then fails discretely, because a wall is a discontinuity
that a smooth latent predictor crosses when it should not. On PushT error
accumulates gradually instead. In both cases the first block --- the part the
controller actually executes at $m{=}1$ --- is the part the world model gets
right, which is a concrete reason the controller is close to flat in $m$ while
still being better at $m{=}1$.

\subsection{Refinement, made visible}

\begin{figure}[tbp]
  \centering
  \includegraphics[width=.86\linewidth]{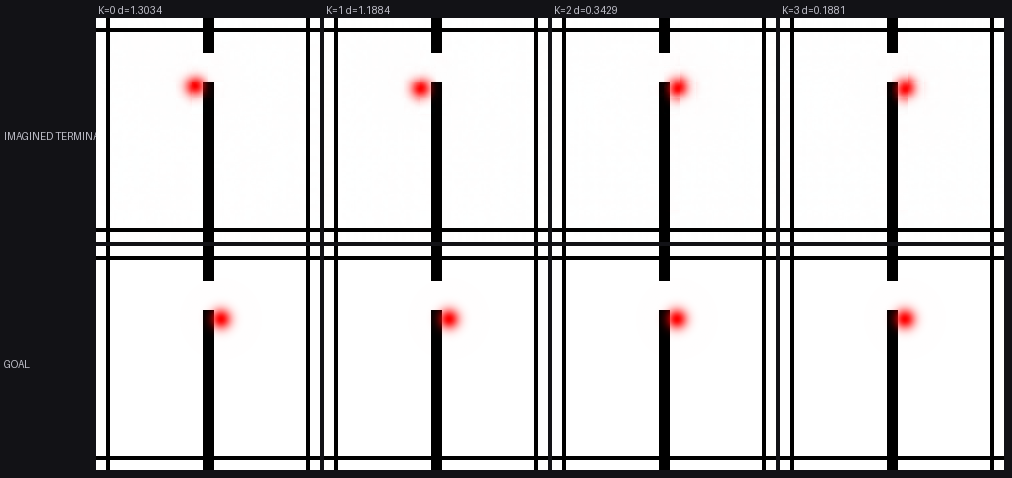}
  \caption{\textbf{What each refinement buys, TwoRooms.} Top: the decoded
  imagined terminal state after refinement $k$. Bottom: the goal. The imagined
  endpoint migrates toward the goal as $k$ increases while the predicted distance
  falls by roughly $7\times$ over three refinements.}
  \label{fig:decoder-refinement}
\end{figure}

\begin{figure}[tbp]
  \centering
  \includegraphics[width=.86\linewidth]{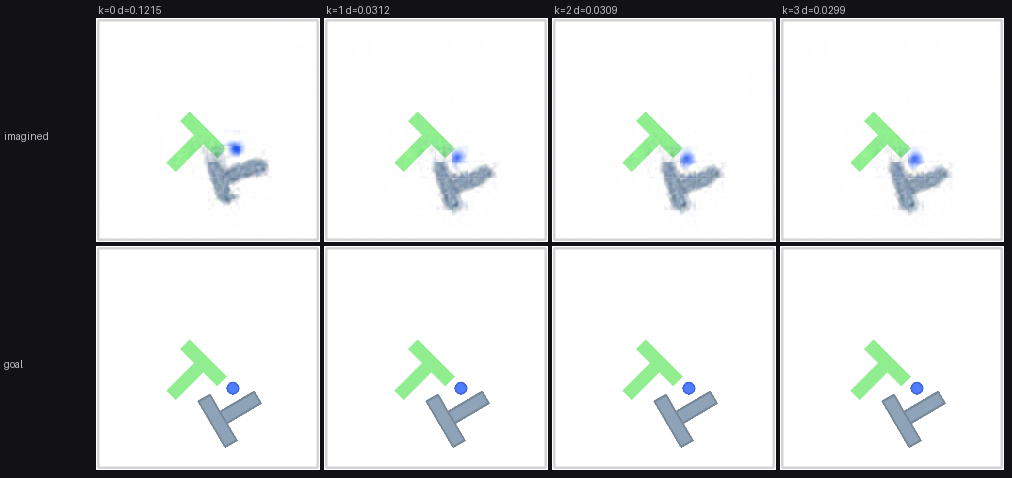}
  \caption{\textbf{What each refinement buys, PushT.} The initial plan $k{=}0$
  is already close, and one refinement captures most of the available
  improvement ($4\times$), after which the plan is essentially settled --- the
  behaviour a converging solver with decaying step sizes should show.}
  \label{fig:decoder-refinement-pusht}
\end{figure}

Figures~\ref{fig:decoder-refinement} and~\ref{fig:decoder-refinement-pusht}
decode the imagined terminal state after each refinement $k=0\ldots K$. On
TwoRooms the predicted goal distance falls $1.3034 \rightarrow 1.1884
\rightarrow 0.3429 \rightarrow 0.1881$; on PushT it falls $0.1215 \rightarrow
0.0312 \rightarrow 0.0309 \rightarrow 0.0299$, with almost all of the gain in the
first refinement. In both the decoded terminal state moves visibly toward the
goal, so the $K$ refinements do real work rather than making cosmetic
adjustments to an already-settled plan. This is the same convergence the learned
step sizes $[0.550, 0.451, 0.312]$ imply, shown in pixels.

\FloatBarrier

\subsection{Executed episodes, end to end}

\begin{figure}[tbp]
  \centering
  \includegraphics[width=\linewidth]{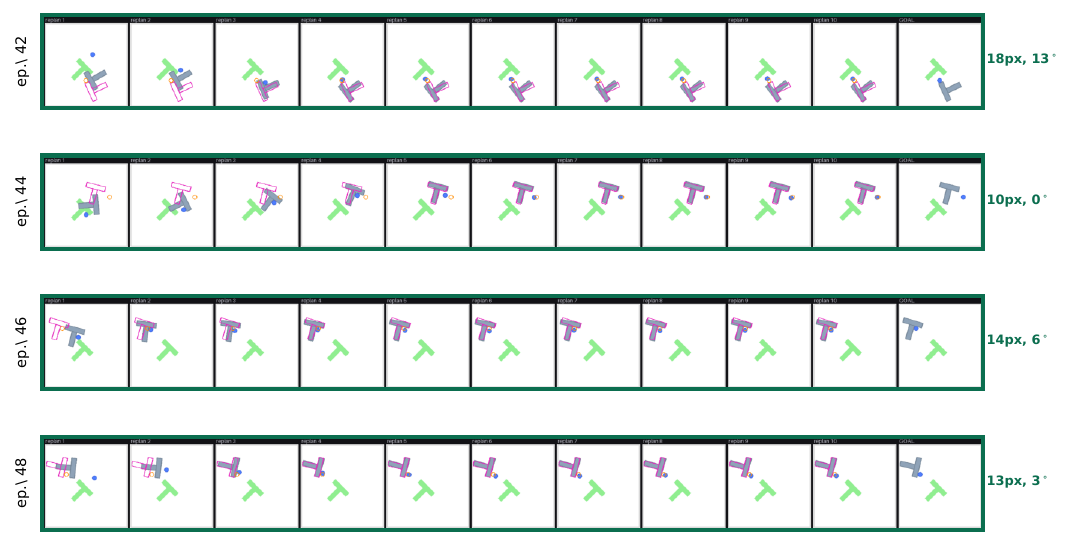}
  \caption{\textbf{PushT: executed rollouts, four successful episodes.} Each row
  is one episode: the state at each of ten replans, then the goal. The magenta
  outline is the target pose, and the final position and orientation error are
  given at the right. Note the first replans in each row, where the pusher
  travels around the block before it can move it. \textbf{These episodes use
  sim-generated goals, not the held-out expert evaluation set that the success
  rates in Section~\ref{sec:results} are computed on}; they illustrate behaviour
  and are not a success rate.}
  \label{fig:exec-pusht}
\end{figure}

\begin{figure}[tbp]
  \centering
  \includegraphics[width=\linewidth]{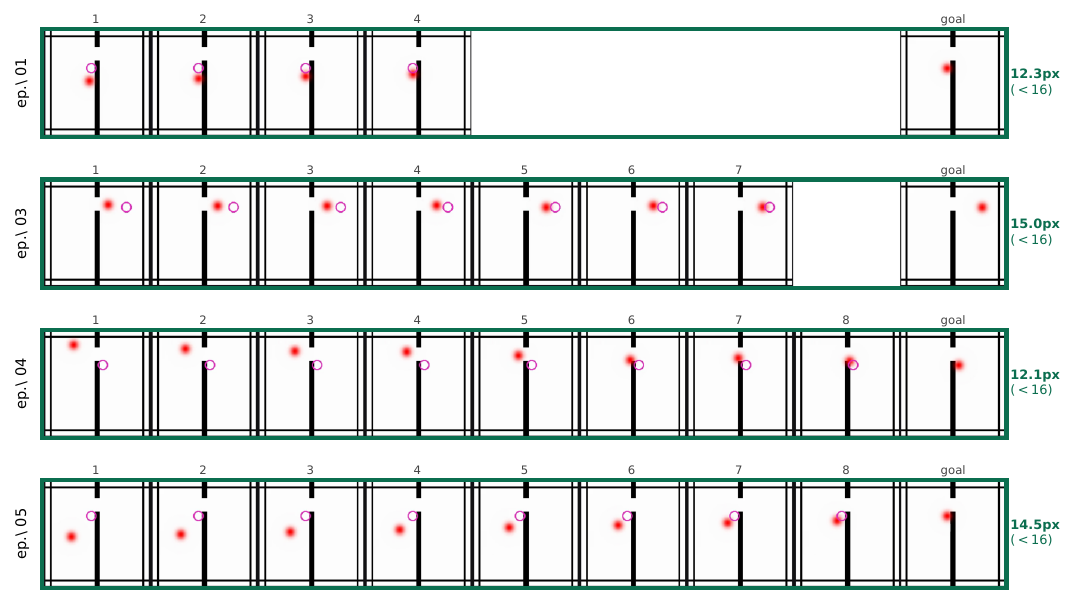}
  \caption{\textbf{TwoRooms: executed rollouts, four successful held-out
  episodes} at $m{=}1$, controller seed $43$. Columns are evenly spaced replan
  points and the last column is the goal; the magenta ring is the $16$-pixel
  success tolerance, and the final physical distance is given at the right.
  Episodes differ in length, so rows with fewer replans are padded with blank
  space rather than repeated frames. These are the same held-out trials used
  throughout Appendix~\ref{sec:decoder}, fixed before inspection.}
  \label{fig:exec-tworooms}
\end{figure}

Figures~\ref{fig:exec-pusht} and~\ref{fig:exec-tworooms} show complete executed
episodes rather than single rollouts: each row is one episode, read left to
right, ending with the goal being scored. The target is drawn as an outline in
every frame, so the reader can see the gap closing rather than having to infer
it. Four episodes are shown per environment, and all of them are successes.

Read together, the two figures show the same closed-loop behaviour in two very
different action spaces. On TwoRooms progress is monotone: the agent moves
toward the doorway, through it, and onto the target, and the replan points are
evenly spaced because nothing interrupts it. On PushT progress is not monotone
in the same way --- the pusher must first travel around the block before any
useful motion happens, so the first two or three replans move the agent without
moving the T at all, and the block only converges once contact is established.
That is the behaviour the arrival term has to permit: a plan that looks like it
is making no progress for two blocks is correct on this task, and an objective
that scored only the immediate step would penalise it.

\subsection{The physical trajectory}

\begin{figure}[tbp]
  \centering
  \includegraphics[width=.46\linewidth]{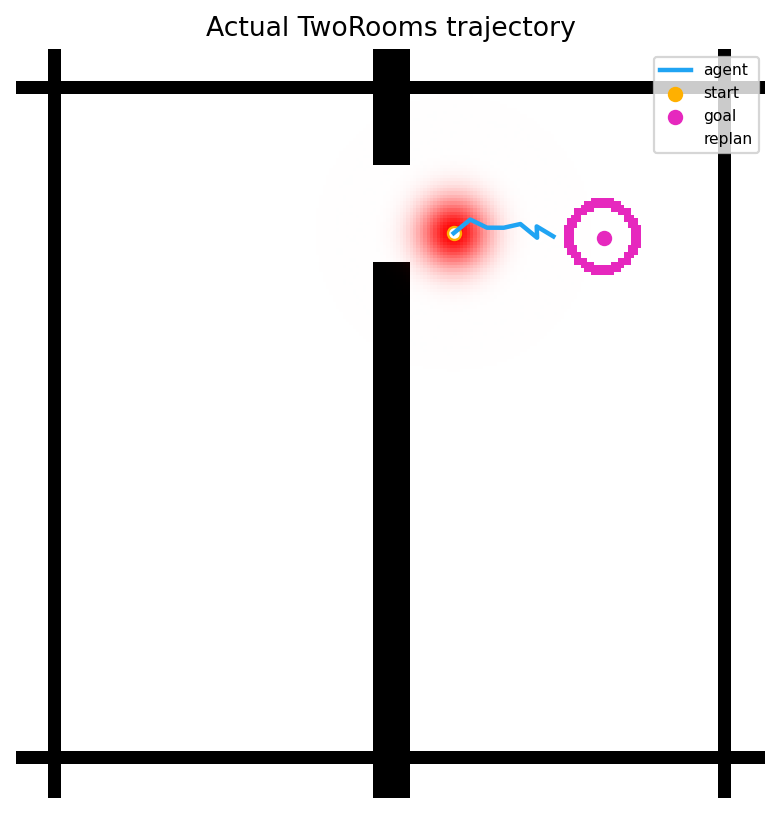}
  \caption{\textbf{Executed TwoRooms trajectory}, $m{=}1$. The agent path is
  drawn over the true floor plan with the goal tolerance ring; physical distance
  falls $44.8 \rightarrow 15.0$ pixels over the episode. Goal markers are drawn
  after rendering and never enter the encoder.}
  \label{fig:decoder-trajectory}
\end{figure}

Figure~\ref{fig:decoder-trajectory} plots an episode in environment
coordinates rather than latents: the real agent path, the true walls, the target
tolerance ring, and the replan points. The trials shown were fixed \emph{before}
any visual inspection, by a stated rule --- the first four manifest trials whose
start-goal separation is at least $16$ pixels, so that a trial cannot be
trivially solved by standing still --- and the selection did not consult
controller outcomes.

\paragraph{Scope.}
We trained decoders for TwoRooms and PushT, and make no decoded claims about
Reacher or Cube. No quantitative result in Section~\ref{sec:results} depends on
any decoder: it is a visualization probe and never enters control or evaluation.

\section{The full algorithm for the architecture: }
\begin{algorithm}[t]
\caption{Iterative latent-space controller}
\label{alg:iterative-controller}
\small
\begin{algorithmic}[1]
\Require Recent observations $o_{t-N+1:t}$, goal observation $o_G$,
past action blocks $b_{t-N+1:t-1}$, frozen encoder $E$,
frozen action-conditioned predictor $P$, horizon $H$,
refinement budget $K$, execution horizon $rh$.
\Ensure Execute the first $rh$ blocks of the final action plan.
\State $x_{t-N+1:t} \gets E(o_{t-N+1:t})$ \Comment{frozen image encoder}
\State $x_G \gets E(o_G)$
\State $C \gets \mathrm{Conditioning}(x_{t-N+1:t}, x_G)$
\Comment{context tokens and one goal token}
\State $Y \gets \mathrm{InitialPlan}(C)$
\Comment{$H$ learned plan tokens; no rollout consequences initially}
\For{$k \gets 0$ to $K$}
    \State $b^{k} \gets \mathrm{ActionHead}(Y)$
    \Comment{$H$ bounded action blocks}
    \State $e \gets P.\mathrm{ActionEncoder}([b_{t-N+1:t-1}; b^{k}])$
    \State $\hat{X} \gets x_{t-N+1:t}$
    \For{$j \gets 1$ to $H$}
        \State $W_j \gets \mathrm{RecentWindow}(\hat{X}, N)$
        \State $A_j \gets \mathrm{AlignedActions}(e, W_j)$
        \State $\hat{x}_j \gets P(W_j, A_j)$ \Comment{frozen one-step prediction}
        \State $\hat{X} \gets [\hat{X}; \hat{x}_j]$
    \EndFor
    \State $\hat{X}_{\mathrm{future}} \gets (\hat{x}_1,\ldots,\hat{x}_H)$
    \Comment{$H$-step imagined rollout}
    \If{$k = K$}
        \State \textbf{break}
    \EndIf
    \For{$j \gets 1$ to $H$}
        \State $\delta_j \gets \hat{x}_j - x_G$
        \State $d_j \gets \mathrm{mean}(\delta_j^2)$
        \State $q_j \gets [Y_j; \hat{x}_j; \delta_j; d_j]$
        \Comment{plan token, outcome, goal error, and distance}
    \EndFor
    \State $R \gets \mathrm{ConsequenceEncoder}(q_{1:H}, C)$
    \Comment{one consequence token per planned block}
    \State $Z \gets \mathrm{JointRefiner}(Y, R, C)$
    \Comment{joint reasoning over plan, context, and goal tokens}
    \State $\Delta Y \gets \mathrm{DeltaHead}(Z)$
    \State $Y \gets Y + \sigma(\eta_k)\Delta Y$
    \Comment{learned gated residual correction}
\EndFor
\State Execute the first $rh$ blocks of $b^{K}$ in the environment
\State Observe new frames, update the history, and repeat
\end{algorithmic}
\end{algorithm}


\end{document}